*This is a draft of a report to which the book chapter is based on.

## Two Decades of Local Binary Patterns – A Survey

Matti Pietikäinen and Guoying Zhao

Abstract

Texture is an important characteristic for many types of images. In recent years very discriminative and computationally efficient local texture descriptors based on local binary patterns (LBP) have been developed, which has led to significant progress in applying texture methods to different problems and applications. Due to this progress, the division between texture descriptors and more generic image or video descriptors has been disappearing. A large number of different variants of LBP have been developed to improve its robustness, and to increase its discriminative power and applicability to different types of problems. In this chapter, the most recent and important variants of LBP in 2D, spatiotemporal, 3D, and 4D domains are surveyed. Interesting new developments of LBP in 1D signal analysis are also considered. Finally, some future challenges for research are presented.

## 12.1. Introduction

Texture is an important characteristic of many types of images. It can be seen in images ranging from multi-spectral remotely sensed data to microscopic images. Texture can play a key role in a wide variety of applications of computer vision and image analysis. Therefore, the analysis of textures has been a topic of intensive research since the 1960s. Most of the proposed methods have not been, however, capable to perform well enough for real-world textures.

In recent years, very discriminative and computationally efficient local texture descriptors have been developed, such as local binary patterns (LBP), which has led to significant progress in applying texture methods to different problems and applications. The focus of research has broadened from 2D textures to 3D textures and spatiotemporal (dynamic) textures. Due to this progress, the division between texture descriptors and more generic image or video descriptors has been disappearing.

The LBP operator can be seen as a unifying approach to the traditionally divergent statistical and structural models of texture analysis. Perhaps the most important property of the LBP operator is its robustness to monotonic gray-scale changes caused, for example, by illumination variations. Another important property is its computational simplicity, which makes it possible to analyze images in challenging real-time settings.

The original LBP was invented already two decades ago, published first at ICPR 1994 and then in Pattern Recognition (Ojala et al, 1996), but at that time one could not imagine what a great success it would be today. In the 1990s LBP did not receive interest in the scientific community, because it was regarded as an ad hoc method. However, the promising power of LBP was already evident, because it performed much better in texture classification and segmentation tasks than the state of the art at that time (Ojala & Pietikäinen, 1999). Due to its computational simplicity and good performance, LBP was also successfully used in some applications such as industrial inspection. The theoretical foundations of LBP became much more clear after our research on multidimensional distributions of signed gray-level differences, carried out jointly with Prof. Erkki Oja and Dr. Kimmo Valkealahti (Ojala et al., 2001).

To the large extent the scientific community accepted LBP after its generalized version was published in IEEE PAMI journal (Ojala et al., 2002), and even more after it was shown to be highly successful in face recognition, published first at ECCV 2004 and then in IEEE PAMI (Ahonen et al., 2006). In 2014, the ECCV paper was awarded with the prestigious Koenderink Prize for Fundamental Contributions in Computer Vision. Different types of applications of LBP to motion analysis have been proposed after the spatiotemporal LBP was introduced, also in IEEE PAMI (Zhao & Pietikäinen, 2007). All these papers are highly cited, reflecting the increasing popularity of LBP. Due to this success of LBP, a book describing the basic methods and surveying different variants and applications was published (Pietikäinen et al., 2011). Another survey on LBP and its applications in face analysis appeared in (Huang et al., 2011). An edited book on some local binary pattern variants and applications was published in 2013 (Brahnam et al., 2013).

After these surveys, which covered the progress until 2010, the interest on LBP has been further growing. LBP is no longer just as a simple texture operator, but it forms the foundation for a new direction of research dealing with local binary image and video descriptors. Many different variants of LBP have been proposed to improve its robustness, and to increase its discriminative power and applicability to different types of problems. A large number of new papers have been published in leading journals and conferences. Due to its discriminative power and computational simplicity, the LBP operator has become highly popular in various applications, including, for example, facial image analysis, biometrics, medical image analysis, motion and activity analysis, and content based retrieval from image and video databases.. Fig. 12.1 depicts the number of citations in Web of Science and Google Scholar to the landmark LBP paper published in 2002 (Ojala et al., 2002), showing a clear increase especially after LBP was successfully adopted for face recognition in 2006 and its spatiotemporal version was proposed in 2007.

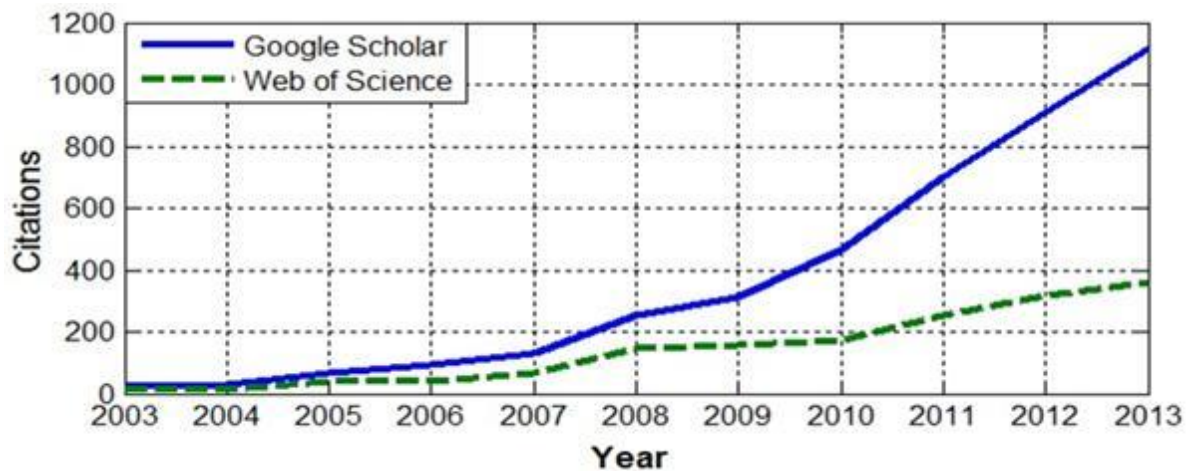

Fig. 12.1. Annual citations to the PAMI 2002 paper on LBP.

Based on all this the publication of this survey is very timely, covering the progress until early 2014. It complements and extends of our preliminary survey presented in Secs. 2.9 and 3.5 of (Pietikäinen et al., 2011), by focusing on the most important recent developments, new types of variants, and future challenges.

12.2. An overview of basic LBP operators

The local binary pattern is based on the assumption that texture has locally two complementary aspects, a pattern and its strength. The pixels of an image are labeled by thresholding the neighborhood of each pixel and the result is considered as a binary number. The distribution of the LBP labels computed over a region is then used for texture description.

The original LBP operator shown in Fig. 12.2 works in a 3 x 3 neighborhood, using the center value as a threshold (Ojala et al., 1996). The thresholded values are multiplied with weights of the corresponding pixels, and summing up the result an LBP code is obtained. The contrast measure C is obtained by subtracting the average gray level of the pixels below the threshold from that of those above (or equal to) the center pixel. If all neighbors of the center pixel have the same value (1 or 0), the value of C is set to 0.

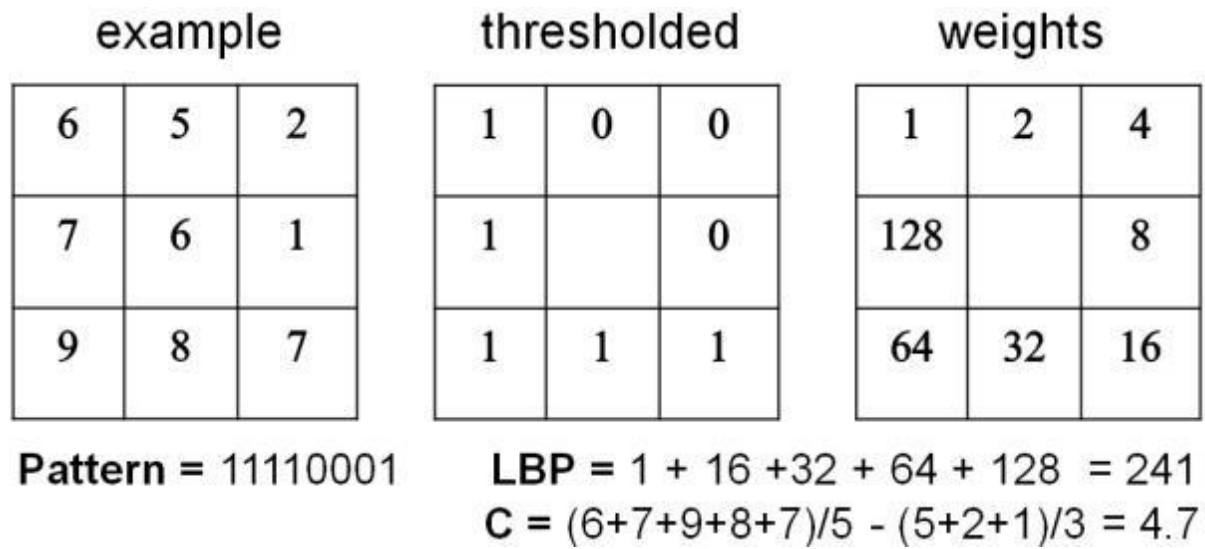

Fig. 12.2. The original LBP.

The distributions of LBP codes or 2D distributions of LBP and C are used as features in texture recognition. The LBP operator was extended to use neighborhoods of different sizes in (Ojala et al., 2002). Using a circular neighborhood and bilinearly interpolating values at non-integer pixel coordinates allow any radius and number of pixels in the neighborhood. The multiscale LBP (MLBP) is obtained by concatenating histograms produced by operators at different radii. The gray scale variance of the local neighborhood can be used as the complementary contrast measure.

Ojala et al. (2002) found that some of the LBP patterns occur more frequently than others, representing e.g. edges, curves, line-ends, flat areas, and spots. Based on this observation so called uniform patterns were defined to reduce the number of patterns. Uniform patterns have been widely used, and were necessary, e.g. to reduce the length of the feature vector in face description.

In their recent paper, Guo et al. (2010) proposed a completed modeling of the LBP operator and developed an associated completed LBP (CLBP) scheme for texture classification. The local differences are decomposed into two complementary components: the signs (like LBP) and the magnitudes (Fig. 12.3). The magnitude component provides an effective alternative for the complementary contrast measure of LBP. In addition to this texture-related information, they also include information about image intensity in their representation.

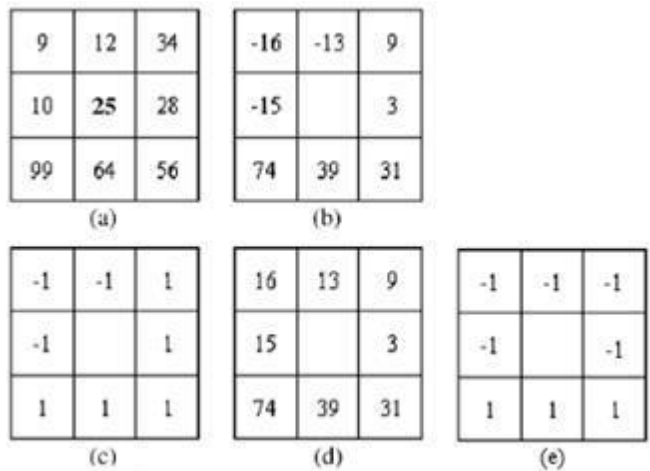

Fig. 12.3. CLBP (a) 3x3 sample block; (b) the local differences; (c) the sign; (d) magnitude components and (e) thresholded binary value of (d) against the average magnitude in whole image region.

Dynamic texture descriptors provide a very effective new tool for motion analysis. Zhao and Pietikäinen (2007) extended the original LBP to a spatiotemporal representation for dynamic texture analysis. For this purpose, the so called Volume Local Binary Pattern (VLBP) operator was proposed, in which dynamic texture is considered as a set of volumes in the (X,Y,T) space where X and Y denote the spatial coordinates and T denotes the frame index (time). The neighborhood of each pixel is thus defined in three dimensional space. The VLBP combines motion and appearance together to describe dynamic textures. To make VLBP computationally simple and easy to extend, an operator based on co-occurrences of local binary patterns on three orthogonal planes (LBP-TOP) considers three orthogonal planes: XY, XT and YT, and concatenates local binary pattern co-occurrence statistics in these three directions as shown in Fig. 12.4. The circular neighborhoods are generalized to elliptical sampling to fit to the space-time statistics.

The first application problems to which spatiotemporal LBP was adopted were facial expression recognition, face and gender recognition, lipreading, and action recognition (Pietikäinen et al., 2011).

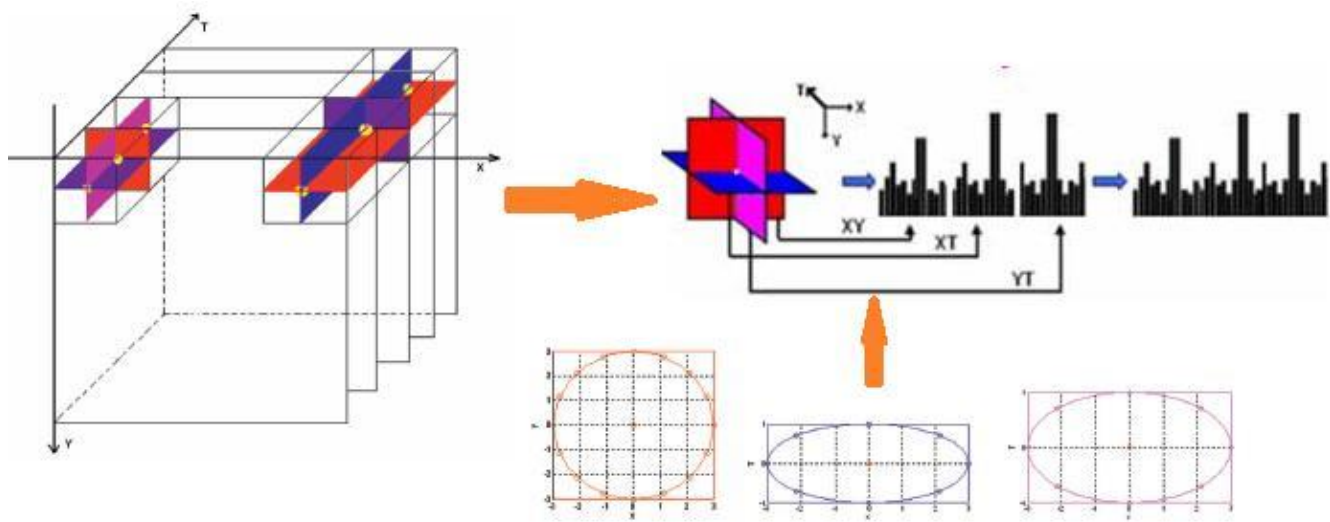

Fig.12.4. LBP-TOP representation for spatiotemporal description.

Facial images are composed local texture patterns, and based of this observation Ahonen et al. (2006) introduced a method for combining local and global information for face (object) description. The face image is divided into several regions from which the LBP feature distributions are extracted and concatenated into an enhanced feature vector to be used as a face descriptor.

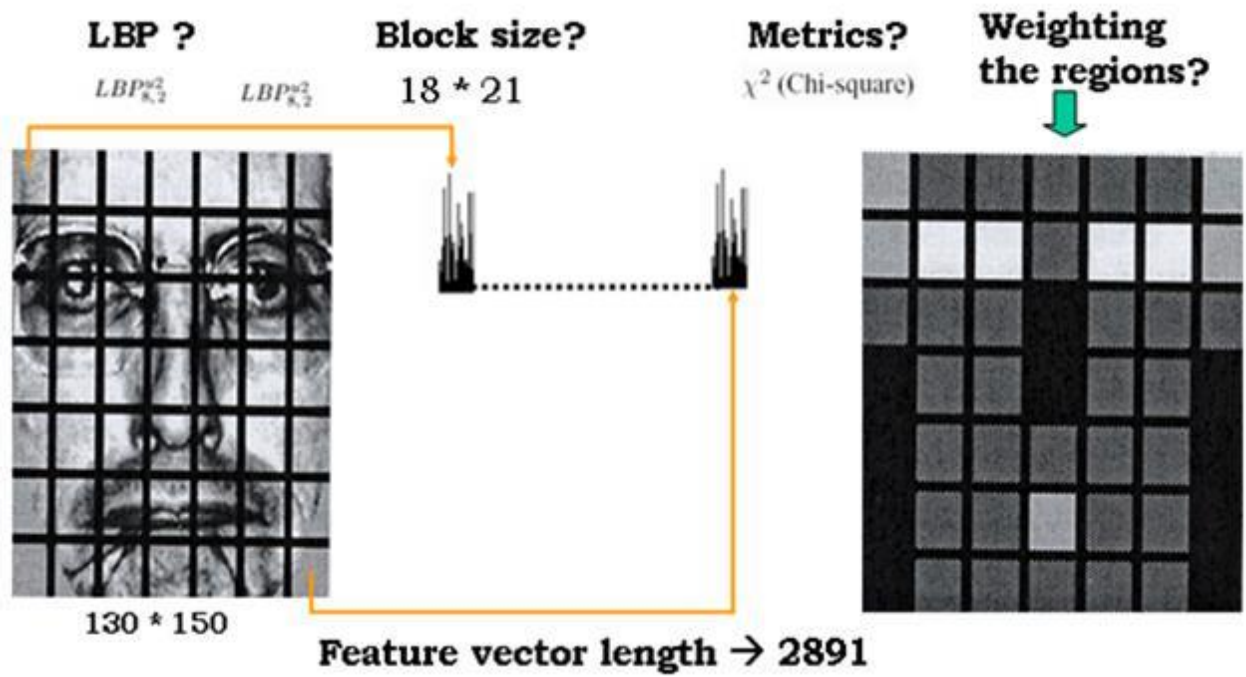

Fig. 12.5. Face description using LBP.

This approach has emerged as one of the key paradigms for facial image analysis, and has also been adopted to other application domains. The length of the feature vector is an important design parameter in this description. Ahonen et al. used uniform patterns to reduce the dimensionality. Fig. 12.5 illustrates the use of LBP for face description (Ahonen et al., 2006).

12.3. LBP variants in spatial domain

*12.3.1 Different types of variants*

Due to its flexibility the LBP method can be easily modified to make it suitable for the needs of different types of problems. Several extensions and modifications of LBP have been proposed with an aim to increase its robustness and discriminative power. In this section different variants are divided into such categories that describe their roles in feature extraction. Some of the variants could belong to more than one category, but only the most obvious categories are here considered. The choice of a proper method for a given application depends on many factors, such as the discriminative power, computational efficiency, robustness to lighting and other variations, and the imaging system used.

Preprocessing

In many applications, it is useful to preprocess the input image prior to LBP feature extraction. Especially multi-scale Gabor filtering and edge detection have been used for this purpose. Gabor filtering has been widely used before LBP computation in face recognition. A motivation for this is that methods based on Gabor filtering and LBP provide complementary information: LBP captures small and fine details, while Gabor filters encode appearance information over a broader range of scales. For example, W. Zhang et al. (2005)

proposed the extraction of LBP features from images obtained by filtering a facial image with 40 Gabor filters of different scales and orientations. The extracted features are called Local Gabor Binary Patterns (LGBP). A downside of the method is the high dimensionality of the LGBP representation.

Tan and Triggs (2010) developed a very effective preprocessing chain for compensating illumination variations in face images. It is composed of gamma correction, difference of Gaussian (DoG) filtering, masking (optional) and equalization of variation. This approach has been very successful in LBP-based face recognition under varying illumination conditions. When using it for the basic LBP, the last step can be omitted due to LBP's invariance to monotonic gray scale changes.

In many recent studies edge detection has been used prior to LBP computation to enhance the gradient information. Gradient images are more insensitive to lighting variations than the original images. Perhaps the first one was Yao and Chen (2003) proposing local edge patterns (LeP) to be used with color features for color texture retrieval. In LeP, the Sobel edge detection and thresholding are used to find strong edges, and then LBP-like computation is used to derive the LeP patterns.

Li et al. (2010) presented an approach based on capturing the intrinsic structural information of face appearances with multi-scale heat kernel matrices. Heat kernels perform well in characterizing the topological structural information of face appearance. Histograms of local binary patterns computed for non-overlapping blocks are then used for face description.

Also other types of preprocessing have been applied with LBP, including wavelets and momentograms, as described in sections Multiscale analysis and Combining local and global information.

Neighborhood topology and sampling

One important factor which makes the LBP approach so flexible to different types of problems is that the topology of the neighborhood from which the LBP features are computed can be different, depending on the needs of the given application.

The extraction of LBP features is usually done in a circular or square neighborhood. A circular neighborhood is important especially for rotation-invariant operators. However, in some applications, such as face recognition, rotation invariance is not required, but anisotropic information may be important. To exploit this, Liao and Chung (2007a) proposed an elliptical binary pattern (EBP) for face recognition. Nanni et al. (2010) investigated the use of different neighborhood topologies (circle, ellipse, parabola, hyperbola and Archimedean spiral) and encodings in their research on LBP variants for medical image analysis. An operator using quinary encoding in an elliptic neighborhood (EQP) provided the best performance. He et al. (2012) used rotation-invariant LBP with elliptic sampling in four directions together with circular sampling to get anisotropic and isotropic information. A Multi-structure LBP (Ms-LBP) was achieved by applying this operator at different layers of an image pyramid. A downside of the method is that quite large image samples are needed for extracted macrostructures. Wolf et al. (2011) considered different ways of using bit strings to encode the similarities between patches of pixels, which could capture complementary information to pixel-based descriptors. They proposed a Three-Patch LBP (TPLBP) and Four-Patch-LBP (FPLBP). For each pixel in TPLBP, for example, a w x w patch centered at the pixel and S additional patches distributed uniformly in a ring of radius r around it are considered. Then, the values for pairs of patches located on the circle at a specified distance apart are compared with those of the central patch. The value of a single bit is set according to which of the two patches is more similar to the central patch. The code produced will have S bits per pixel. In FPLBP, two rings centered on the pixel were used instead of one ring in TPLBP.

Orjuela et al. (2013) presented Geometrical Local Textural Patterns (GLTP), which are based on exploring intensity changes on oriented neighborhoods. An oriented neighborhood describes a particular geometry composed of points on circles with different radii around the center pixel. A digital representation of the points on the oriented neighborhood defines a GLTP code. The simple case called Geometric Local Binary Pattern is based on Boolean comparisons.

Wang et al. (2013) proposed a sampling structure based on combining pixel and patch to mimic the retinal sampling grid (Pixel to Patch, PTP). Also a neighboring intensity relationship (NIR) operator was proposed to complement LBP texture information by exploring gray-scale properties between neighborhoods. Two rotation invariant descriptors were also proposed: The local intensity relationship pattern (LNIRP) based on the NIR operator and LNIRP_PTP using the PTP sampling structure. The method is computationally simple, has small feature dimensionality, and is training-free. Ylioinas et al. (2013a) introduced a dense sampling approach through the form of up-sampling to extract more stable and discriminative texture patterns in local regions. Experiments on face recognition, texture classification and age group estimation problems on various challenging benchmark databases demonstrate the efficiency of the proposed scheme.

Thresholding and encoding

A drawback of the LBP method, as well as of all local descriptors that apply vector quantization, is that they are not robust in the sense that a small change in the input image would always cause a small change in the output. LBP may not work properly for noisy images or on flat image areas of constant gray level. This is due to the thresholding scheme of the operator
Instead of using the value of the center pixel for thresholding in the local neighborhood, other techniques have also been considered. Hafiane et al. (2007) proposed Median Binary Pattern (MBP) operator by thresholding the local pixel values, including the center pixel, against the median within the neighborhood. The so-called Improved LBP, on the other hand, compares the values of the neighboring pixels against the mean gray level of the local neighborhood (Jin et al., 2004).

Tan and Triggs (2010) proposed a three-level operator called local ternary patterns (LTP), (using one threshold T) e.g. to deal with problems on near constant image areas. In ternary encoding the difference between the center pixel and a neighboring pixel is encoded by three values (1, 0 or -1) according to a threshold T. The ternary pattern is divided into two binary patterns taking into account its positive and negative components. The histograms from these components computed over a region are then concatenated. Fig. 12.6 depicts an example of splitting a ternary code into positive and negative codes.

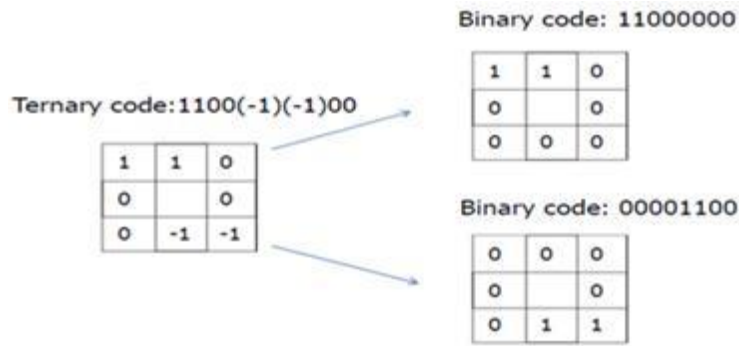

Fig. 12.6. Local ternary pattern (LTP) operator.

In order to make the LBP more robust against negligible changes in pixel values, the thresholding scheme of the operator was modified in (Heikkilä & Pietikäinen, 2006) by adding a small constant $a$ in local difference computations. The bigger the value of $|a|$ is, the bigger changes in pixel values are allowed without affecting the thresholding results. In order to retain the discriminative power of the LBP operator, a relatively small value should be used. An advantage of this modified LBP compared to the three-valued LBPs described above is that the feature vector length remains the same as in the ordinary LBP. Heikkilä et al. (2009) proposed Center-Symmetric LBP (CS-LBP) to reduce the feature vector length of the original LBP. The center-symmetric pairs of pixels in the neighborhood are compared, instead of comparing all neighbors with the center pixel. Therefore the CS-LBP produces only 16 patterns instead of 256 for the basic LBP. An interest region descriptor based on CS-LBP compared favorably to the SIFT operator in image matching and categorization experiments, especially for images with lighting variations. A similar thresholding approach as in (Heikkilä & Pietikäinen, 2006) was used to improve the robustness of CS-LBP descriptor.

Nanni et al. (2010) studied the effects of different encodings of the local gray-scale differences, using binary (B), ternary (T) and a quinary (Q) encodings. In binary coding, the difference between a neighboring pixel and the center pixel is encoded by two values (0 and 1) like in LBP, in ternary encoding it is encoded by three values as in LTP, and in quinary encoding by five values (-2.-1,0,1,2) according to two thresholds (T1 and T2). A quinary code can be split into four binary LBP codes.

A downside of the methods using one or two thresholds is that the methods are not strictly invariant to local monotonic gray level changes as the original LBP. The feature vector lengths of these operators are also longer.

A soft three-valued LBP using fuzzy membership functions was proposed by Ahonen and Pietikäinen (2007) to improve the robustness. In soft LBP, one pixel typically contributes to more than one bin in the histogram. An extensive study on generalized fuzzy LBPs was carried out by Katsigiannis et al. (2013). A disadvantage of the fuzzy methods is their increased computational cost.

Liao et al. (2010) noticed that adding the small threshold (T) in LTP is not invariant under scaling of intensity values. The intensity scale invariant property of a local comparison operator is very important, e.g., in background modeling, because lighting variations, either global or local, often cause sudden changes of gray scale intensities of neighboring pixels simultaneously, which would approximately be a scale transform with a constant factor. Therefore, a Scale Invariant Local Ternary Pattern (SILTP) operator was developed for dealing with the gray scale intensity changes in complex background. They also proposed a pattern kernel density estimation technique to effectively model the probability distribution of local patterns in the pixel process. Ylioinas et al. (2013b) adopted a related method for image appearance description. Compared to the standard LBP histogram statistics where one labeled pixel always contributes to one bin of the histogram, the proposed method exploits a kernel-like similarity function to determine weighted votes contributing several possible pattern types in the statistic. As a result, the method yields a more reliable estimate of the underlying LBP distribution of the given image, providing improved performance especially for small-sized samples.

Trefny and Matas (2010) proposed two encoding schemes, which are complementary to the standard LBPs and also invariant to monotonic intensity transformations. The binary value transition coded LBP (tLBP) is composed of neighbor pixel comparisons in clockwise direction for all pixels except the central, encoding relation between neighboring pixels. Direction coded LBP (dLBP) is related to CS-LBP operator, but uses also center pixel information for encoding. Intensity variation along each of the four basic directions is coded into two bits. The first bit encodes whether the center pixel is an extreme point and the second bit encodes whether difference of border pixels compared to the center pixel grows or falls.

Liu et al. (2012) extracted two different and complementary types of features, pixel intensities and differences, from local patches. The intensity-based features consider the intensities of the central pixel (CI) and its neighbors (NI); while for the difference-based feature, two components are computed: the radial-difference (RD) and the angular-difference (AD). Two intensity-based descriptors CI-LBP and NI-LBP, and two difference-based descriptors RD-LBP and AD-LBP were proposed.

Inspired by LBP, higher order local derivative patterns (LDP) were proposed by B. Zhang et al. (2010), with applications in face recognition. The basic LBP represents the first-order circular derivative pattern of images, a micropattern generated by the concatenation of the binary gradient directions as was shown in (Ahonen & Pietikäinen, 2009). The higher order derivative patterns extracted by LDP will provide more detailed information, but may also be more sensitive to noise than in LBP. The length of the feature vector is also four times the length of LBP and the issue of rotation invariance was not addressed. To address these problems, Z. Guo et al. (2012) proposed local directional derivative patterns (LDDP), a special case of which the LBP is. Like for LBP, rotation invariance can be easily defined also for LDDP. Best recognition accuracy can be obtained by combining 1st and 2nd order LDDPs.

Hussain and Triggs (2012) tackled the problem of earlier LBP variants based on hand-specific codings, which limits them to small spatial support and coarse gray level comparisons. They proposed Local Quantized Patterns (LQP), using vector quantization to code larger or deeper patterns. Precomputed lookup tables are used to make coding very fast. Their approach outperformed HOG, LBP and LTP in challenging object detection and texture classification problems. Later, Huang et al. (2013) introduced CLQP to include also magnitude (like in CLBP) and orientation information. A better and much faster way for initializing the vector quantization was also proposed.

Jiang et al. (2012) proposed gradient LBPs (GLBP) for human detection. After LBP is calculated for each pixel, a 56 bin histogram is built with the co-occurrence of width value and angle value of LBPs. Width is the number of value 1s in the binary code of this pixel. Eight direction codes with 0 to 7 are defined in the direction of eight neighbor pixels. Angle value is the direction code of the middle pixel in 1 area of its binary code. Magnitude of gradient is used as weight to this 56 bins histogram.

Multiscale analysis

From a signal processing point of view, the sparse sampling used by multiscale LBP operator (MLBP) (Ojala et al., 2002) may not result in an adequate representation of the signal, resulting in aliasing effects. Due to this some low-pass filtering would be needed to make the operator more robust. From the statistical point of view, however, even sparse sampling is acceptable provided that the number of samples is large enough. The sparse sampling is commonly used for example with the methods based on gray scale difference or co-occurrence statistics. Mäenpää and Pietikäinen (2003) proposed to use Gaussian low-pass filters for collecting texture information from a larger area than the original single pixel. The filters and sampling position were designed to cope with the neighborhood as well as possible while minimizing the redundant information. With this approach, the radii of the LBP operators used in the multiscale version grow exponentially.

Another extension of multiscale LBP operator is the multiscale block local binary pattern (MB-LBP) (S. Liao et al., 2007) which has gained popularity especially in facial image analysis. The key idea of MB-LBP is to compare average pixel values within small blocks instead of comparing pixel values. Instead of the fixed uniform pattern mapping, MB-LBP was proposed to be used with a mapping that is dynamically learned from a training data. In this mapping, the N most often occurring MB-LBP patterns receive labels 0…N-1, and all the remaining patterns share a single label. The number of labels, and consequently the length of the MB-LBP histogram is a parameter the user can set.

A straightforward way for multiscale analysis is to utilize a pyramid of the input image computed at different resolutions, and then concatenate LBP distributions computed from different levels of the pyramid. In their research on contextual analysis of textured scene images Turtinen and Pietikäinen (2006) combined this kind of idea with the original multiscale LBP approach: image patches at three different scales were resized to the same size and then LBP features were computed using LBPs with three different radii.

He et al. (2010) developed a pyramid-based multistructure LBP for texture classification. It is obtained by executing the LBP on different layers of image pyramid, allowing to extract both micro and macro structures from textures. Five templates are used for creating the pyramid. The first one is a 2D Gaussian function used to smooth the image. Other four anisotropic filters are used to create anisotropic subimages of the pyramid in four directions. Qian et al. (2011) also extended LBP to pyramid transform domain. Different ways of sampling in pyramid construction are considered, including no sampling, partial sampling and spatial pyramid sampling, of which the last one leads to the smallest computational cost.

Song and Li (2013) combined wavelets and LBP. They build up the image description using a hierarchical framework based on low-dimensional WaveLBP features, which not only extracts multi-scale oriented features and local image patterns, but also captures multi-level (pixel, patch, image) features. A very competitive performance was demonstrated in experiments.

Liu et al. (2014) proposed a computationally simple approach to multi-scale analysis with their Binary Rotation Invariant and Noise Tolerant Texture descriptor (BRINT). Points are sampled in a circular neighborhood, but the number of bins in a single scale LBP histogram is kept constant and small by averaging over several contiguous pixels in the circle before binarization. This allows to encode a large number of scales and also reduces the effects of noise. Both sign and magnitude components, like in CLBP, are considered.

Handling rotation and scale variations

LBPs have been used for rotation invariant texture recognition since late 1990s (Pietikäinen et al., 2000). In the most widely used version proposed in (Ojala et al., 2002), the neighboring n binary bits around a pixel are clockwise rotated n times. A maximal number of the most significant bits is used to express this pixel.

Z. Guo et al. (2010b) developed an adaptive LBP (ALBP) by incorporating the directional statistical information for rotation invariant classification. The directional statistical features, specifically the mean and standard deviation of the local absolute difference are extracted and used to improve the LBP classification efficiency. In addition, the least square estimation is used to adaptively minimize the local difference for more stable directional statistical features. Garcia et al. (2013) extended the ALBP method to an approach called adaptive local binary pattern with oriented standard deviation (ALBPS), which adds an oriented standard deviation term to the LBP operator instead of using this information in the matching. Excellent results were obtained when assessing boar sperm vitality.

In (Z. Guo et al., 2010c), LBP variance (LBPV) was proposed as a rotation invariant descriptor. For LBPV there are three stages: 1) Using the local contrast (gray scale variance) as an adaptive weight to adjust the contribution of the LBP code in histogram calculation. 2) Learning the principal directions. The extracted LBPV features are used to estimate the principal orientations, and then the features are aligned to the principal orientations, and 3) Determining the non-dominant patterns and thus by reducing them, feature dimension reduction was achieved.

G. Zhang et al. (2010) proposed Monogenic-LBP (M-LBP), which integrates the traditional rotation-invariant LBP operator with two other rotation-invariant measures: the local phase and local surface type computed by the first and second order Riesz transforms, respectively. The local phase corresponds to a qualitative measure of local structure (step, peak etc.), whereas the monogenic curvature tensor extracts local surface type information. Y. Zhao et al. (2012) extracted grayscale difference information, but totally abandoned structural information by proposing a completed local binary count (CLBC) operator. Their results suggest that microstructure is not always needed for rotation invariant classification.

Zhao et al. (2012) proposed an approach to compute rotation invariant features from histograms of local, non-invariant patterns. They applied this approach to both static and dynamic LBP descriptors. For static textures, they presented Local Binary Pattern Histogram Fourier features (LBP-HF). LBP-HF is computed from discrete Fourier transforms of LBP histograms. The approach can also be generalized to embed any uniform features into this framework, and combining supplementary information, e.g. sign and magnitude components of LBP together can improve the description ability. Moreover, two variants of rotation invariant LBP-HF descriptors were proposed for LBP-TOP, which is not rotation invariant. Experiments showed that LBP-HF and its extensions perform very well in rotation invariant texture classification. They are also robust with respect to changes in viewpoint, outperforming recent methods proposed for view-invariant recognition of dynamic textures.

Li et al. (2012) considered scale- and rotation invariant LBP. A circular neighboring set of a pixel is defined as a self-adaptive texton. The optimal scale of each pixel is adaptively selected based on the scale space derived by the Laplacian of Gaussian, and used to determine the radius of the scale-adaptive texton. Different pixels have different optimal scales, resulting in scale invariance. The subuniform patterns of each uniform pattern are defined to improve the discrimination, and the circular shift LBP histogram is computed to obtain rotation invariance.

Considering co-occurrences

Recent studies show that encoding co-occurrences of local binary patterns can significantly improve the performance. Co-occurrences can be considered in different ways, within the same LBP operator, between adjacent operators, or at region level.

Qi et al. (2014) introduced a pairwise rotation invariant co-occurrence LBP (PRI-CoLBP), which incorporates two types of context: spatial co-occurrence and orientation co-occurrence. The method aims to preserve the relative angle between the orientations of individual features. The relative angle provides information about local curvature. For each co-occurrence pattern, the gradient magnitude of the two points is used to weight the co-pattern. Excellent results using a three-scale PRI-CoLBP are reported for many different datasets. The length of the feature vector (590 for the single scale operator and 6 x 590 when extracting six co-occurrence patterns in two angles and three scales) may limit the applicability of this and also many other co-occurrence methods in certain applications, such as face recognition. Qi et al. (2013a) also proposed a rotation-invariant Multi-scale Joint Encoding of LBPs (MSJ-LBP) operator . In the original multi-scale LBP each scale is encoded into histograms separately, and then the histograms are concatenated. This ignores the correlation between different scales. MSJ-LBP encodes jointly the local binary patterns of two scales around the center point to capture their correlation. For each point at two chosen scales, its joint MS-LBP pattern (one of the 590 patterns) and its gradient magnitude are computed. The gradient magnitude is used to weight the given joint pattern.

Nosaka et al. (2011) proposed an Adjacent LBPs (CoALBP) operator. The co-occurrence of adjacent LBPs is defined as an index how often their combination occurs in the whole image (or region). Co-occurrence is measured with an autocorrelation matrix generated from multiple LBPs. Later, they extended it to Rotation Invariant Co-occurrence among adjacent LBPs (RIC-LBP) operator, which is enabled by introducing the concept of rotation equivalence class to CoALBP (Nosaka et al., 2013). Louis and Plataniotis (2011) consider co-occurrences of rotation-invariant LBP patterns (CoLBP) at region level. Basic rotation-invariant LBPs are computed for all possible scales in the examined scanning window. Then, instead of computing histograms, multiple instances of rotation-invariant LBPs are selected using the sequential forward selection algorithm.

Handling color

LBP has also been widely applied to color images. To describe color and texture jointly, opponent color LBP (OCLBP) was defined in (Mäenpää & Pietikäinen, 2004). In opponent color LBP, the operator is used on each color channel independently, and then for pairs of color channels so that the center pixel is taken from one channel and the neighboring pixels from the other. Opposing pairs, such as R-G and G-R are highly redundant, so either of them can be used in the analysis. In total, six histograms (out of nine) are utilized (R, G, B, R-G, R-B, G-B), making the descriptor six times longer than the monochrome LBP histogram. The OCLBP descriptor fares well in comparison to other color texture descriptors. It has been later used successfully, e.g. for face recognition. However, the authors of this paper did not recommend joint color and texture description as in their experiments "all joint color texture descriptors and all methods of combining color and texture on a higher level are outperformed by either color or gray-scale texture alone".

A popular way is to apply the ordinary LBP to different color channels separately. Instead of the original R, G and B channels, other more discriminative and invariant color features derived from them can be used for LBP feature extraction as well. Along this line, Zhu et al. (2010) proposed multiscale color LBPs for visual object classes recognition. Six operators were defined by applying multiscale LBP on different types of channels and then concatenating the results together. >From these the Hue-LBP (computed from the hue channel of the HSV color space), Opponent-LBP (computed over all three channels of the opponent color space) and an Opponent-LBP (computed over two channels of the normalized opponent space) provided good performance in experiments. Later Zhu et al. (2013) extended their approach to image region description using Orthogonal Combination of LBPs (OC-LBP) and six new local descriptors based on OC-LBP enhanced with color information. OC-LBP is obtained by combining the histograms of $P/4$ different 4-orthogonal-neighbor operators. The dimension of the descriptor is $4 \times P$, which is linear with the number of neighboring pixels in comparison to 2P for the original LBP. Color OC-LBP descriptor is obtained by concatenating OC-LBP histograms computed over selected color channels. Experiments on image matching, object recognition and scene classification were used to show the effectiveness of the approach.

Qi et al. (2013b) presented an approach to encode cross-channel texture correlation for color texture classification. The texture correlation between different RGB channels is first empirically studied using LBP as texture descriptor and Shannon's information entropy as correlation measurement. For color texture description, pairwise color channels are jointly encoded to obtain Cross-Channel LBPs (CCLBP). A Multi-Scale CCLBP is also developed. Excellent results were reported.

Banerji et al. (2013) present new image descriptors based on color, texture, shape, and wavelets for object and scene image classification. A three dimensional 3D-LBP descriptor is proposed for encoding both color and texture information, and H-descriptor to integrate the 3D-LBP and the HOG of its wavelet transform, to encode color, texture, shape, and local information. The H-descriptor is comparatively assessed on seven well known color spaces. A new H-fusion is also presented by fusing the PCA features of the H-descriptors in the seven color spaces. Experimental results on the Caltech 256 object categories, the UIUC Sports Event, and the MIT Scene datasets demonstrate excellent performance.

Handling noise

Sensitivity to noise in images is one of the key problems of the original LBP. Kylberg and Sintorn (2013) evaluated the noise robustness of eight LBP variants on five different datasets. None of the descriptors was generally more noise robust for all datasets and noise levels. However, LTP was often among the best performing descriptors and Improved LBP (Jin et al., 2004) often performed slightly better than the original LBP. Median LBP (Hafiane et al., 2007) and Local Quinary Pattern (Nanni et al., 2010) performed worst.

Chen et al. (2013a) proposed a simple robust version of LBP (i.e., RLBP) by changing the coding of bits of LBP, which could otherwise be changed by noise. Experimental results on texture datasets demonstrated that the RLBP outperforms many widely used descriptors and other variants of LBP, especially when noise is added in the images. Experimental results in face recognition also provided very promising results. Ren et al. (2013a) introduced a noise-resistant LBP (NRLBP) aiming to preserve the local image structures in the presence of noise. A small pixel difference is vulnerable to noise, and thus it is first encoded as uncertain, and then its value is determined based on the other bits of LBP code to form a code of local uniform pattern. They also proposed extended noise resistant LBP (ENRLBP) to capture line patterns. Experiments with added Gaussian and uniform noise on various datasets demonstrate the efficiency of their approach. Y. Zhao et al. (2013) proposed a Completed Robust Local Binary Pattern (CTLBP), in which each center pixel is replaced with the average of the neighborhood, as in Improved LBP. To make the operator more robust and stable, a weighted local gray level is also introduced to replace the value of the center pixel. The reported results for the UIUC database are clearly worse than those of Chen et al. (2013a).

Combining local and global information

LBP reflects the correlation among pixels within a local area (e.g., 3 x 3 area for $LBP_{8,1}$), which mainly represents the local information. Recently, there have been many works combining more global or more local information with LBP, for getting a more discriminative description from different feature levels.

Liao and Chung (2007b) extracted first dominant patterns, so called advanced LBPs from images, and labeled their locations with "1", and "0" otherwise. The Gray Level Aura Matrix was used to extract the spatial information from each binary image. Guo et al. (2011) investigated rotation invariant image description with a linear model based descriptor named MiC, which is suited to modeling microscopic configuration of images. To explore multi-channel discriminative information on both the microscopic configuration and local structures, the feature extraction process is formulated as an unsupervised framework. It consists of: 1) the configuration model to encode image microscopic configuration; and 2) local patterns to describe local structural information. In this way, images are represented by a novel feature: local configuration pattern (LCP). Khellah (2011) proposed a method, which computes global rotation invariant features from estimated dominant neighborhood structure and combines them with local LBP features, providing very good performance also in experiments with additive Gaussian noise.

Covariance Matrices (CovMs) capture correlation among elementary features of pixels over an image region. Ordinary LBP features cannot be used as elementary features, since they are not numerical variables in Euclidean spaces. To address this problem, Hong et al. (2014) developed a powerful descriptor, named COV-LBP. Firstly, a variant of LBPs in Euclidean spaces, named the LBP Difference feature (LBPD), was proposed. LBPD reflects how far one LBP lies from the LBP mean of a given image region. Secondly, by applying LBPD together with some other features provided a bank of discriminative features for CovMs, providing very good performance in experiments. Papagostas et al. (2013) introduced Moment-based LBPs. Their approach consists of two steps: the momentogram construction and the application of LBP on it. As a result an enhanced LBP histogram is obtained, which is invariant under common geometric transformations (translation, rotation, scaling), enclosing local as well as global information.

Complementary descriptors

 A current trend in the development of new effective local image and video descriptors is to combine the strengths of complementary descriptors. From the beginning the LBP operator was designed as a complementary measure of local image contrast. An interesting alternative for putting the local contrast into the one-dimensional LBP histogram was proposed by Guo et al. (2010c) in their LBPV method. As presented earlier, Guo et al. (2010a) also introduced the CLBP operator, in which the magnitude component is used as an improved contrast measure. Magnitude-LBP and LBPV both contain supplementary information to LBP. They were embedded to the histogram Fourier framework (Zhao et al., 2012) and concatenated to LBP-HF features as complementary descriptors to improve the description power for dealing with rotation variations.

In addition to applying LBP to Gabor-filtered face images, the joint use of LBP and Gabor, and LBP and LPQ, has provided excellent results in face recognition Tan & Triggs, 2010, Chan et al., 2013). The HOG-LBP, combining LBP with the Histogram of Oriented Gradients operator (Dalal & Triggs, 2005), has performed very well in human detection with partial occlusion handling (Wang et al., 2009). Combining ideas from Haar and LBP features have given excellent results in accurate and illumination invariant face detection (Yan et al., 2008). A CS-LBP method for combining the strengths of SIFT and LBP in interest region description has also been developed (Heikkilä et al., 2009).

The Weber Law Descriptor (WLD) is based on the fact that human perception of a pattern depends not only on the change of a stimulus (such as sound, lighting) but also on the original intensity of the stimulus (Chen et al., 2010). WLD consists of two components: differential excitation and orientation. The differential excitation component is a function of the ratio between two terms: one is the relative intensity differences of a current pixel against its neighbors; the other is the intensity of the current pixel. The orientation component is the gradient orientation of the current pixel. For a given image, the two components are used to construct a concatenated WLD histogram. Joint use of LBP and the excitation component of WLD descriptor, together with the histogram of optic flow in dynamic texture segmentation was considered by Chen et al. (2013) This indicates that the excitation component could be useful in replacing the contrast measure of LBP also in other problems. The method of Liu et al. (2013) also uses differential excitation of WLD together with LBP. The former is improved by using Laplacian of Gaussian.

Jun et al. (2013) proposed local gradient patterns (LGP) and binary histograms of oriented gradients (BHOG), and a hybrid feature which combines several local transform features using AdaBoost. Excellent results were reported in face and human detection experiments. In Nguyen et al. (2013), contour templates representing object shape are used to derive a set of ("edge-like") key points at which local appearance features are extracted. Both spatial and orientation information are used in their computation. At each keypoint non-redundant local binary pattern (NR-LBP) is computed. An object descriptor is formed by concatenating NR-LBP features from all keypoints. Very good performance is obtained for MIT and INRIA datasets. A downside of the method is that computation is quite time-consuming. The paper by Ma et al. (2013) demonstrates that orientation information is critical in human detection. They proposed a Oriented Local Binary Patterns (OLBP) feature, which integrates pixel intensity difference information with texture orientation information to capture salient object feature. Also a set of Edge Orientation Histograms. (EOH) and OLBP based intra-block and inter-block features is proposed to describe cell-level and block-level information. Experiments on INRIA and Caltech datasets demonstrate that the approach has a competitive performance and higher speed than existing detectors.

*12.3.2 Feature selection and learning*

It has been shown by many studies that the dimensionality of the LBP distribution can be effectively decreased by reducing the number of neighboring pixels or by selecting a subset of bins available. In many cases a properly chosen subset of LBP patterns can perform better than the whole set of patterns.

Rule-based selection

Already the early studies on LBP indicated that in some problems considering only four neighbors of the center pixel (i.e. 16 bins) can provide almost as good results as eight neighbors (256 bins). Mäenpää et al. (2000) showed that a major part of the discriminative power lies in a small properly selected subset of patterns. In addition to the uniform patterns they also considered a method based on beam search in which, starting from one, the size of the pattern set is iteratively increased up to a specified dimension D, and the best B pattern sets produced so far are always considered.

Smith and Windeatt (2010) used the fast correlation-based filtering (FCBF) algorithm to select the most discriminative LBP patterns. FCBF operates by repeatedly choosing the feature that is most correlated with a given class (e.g. person identity in case of face recognition), excluding those features already chosen or rejected, and rejecting any features that are more correlated with it than with the class. As a measure of correlation, the information-theoretic concept of symmetric uncertainty is used. When applied to the LBP features, FCBF reduced their number from 107,000 down to 120.

Liao et al. (2009) introduced dominant local binary patterns (DLBP) which make use of the most frequently occurred patterns of LBP to improve the recognition accuracy compared to the original uniform patterns. The method has also rotation invariant characteristics. To obtain discriminative patterns, Y. Guo et al. (2012) presented a learning model which is formulated into a three-layered model. It estimates the optimal pattern subset of interest by simultaneously considering the robustness, discriminative power and representation capability of features. This model is generalized and can be integrated with existing LBP variants such as conventional LBP, rotation invariant patterns, local patterns with anisotropic structure, completed local binary pattern (CLBP) and local ternary pattern (LTP) to derive new image features for texture classification.

Boosting

Boosting has become a very popular approach for feature selection. It has been widely adopted for LBP feature selection in various tasks, e.g. 3D face recognition and face detection. AdaBoost is commonly used for selecting optimal LBP settings (such as the size and the location of local regions, the number of neighboring pixels etc.) or for selecting the most discriminative bins of an LBP histogram. For instance, G. Zhang et al. (2005) used AdaBoost learning for selecting an optimal set for local regions and their weights for face recognition. Since then, many related approaches have been used at region level for LBP-based face analysis. Shan and Gritti (2008), on the other hand, used AdaBoost for learning discriminative LBP histogram bins, with an application to facial expression recognition.

Heng et al. (2012) presented a "shrink boost" method for selecting features from multiple LBP histograms. Motivation for it was that feature selection from sparse and high dimension features using conventional greedy based boosting leads to poor generalization. The shrink boost method solves sparse regularization problem with two iterative steps. A "boosting" step first uses weighted training samples to learn a full high dimensional classifier on all features, avoiding overfitting to few features and improves generalization. Then a "shrinkage" step shrinks least discriminative classifier dimension to zero to remove the redundant features. In well-known (INRIA human detection, DaimlerChrysler pedestrian detection, Bird detection) object detection problems they used "shrink boost" to select sparse features from concatenated LBP histograms of multiple quantization and image channels to learn classifier of additive lookup tables, obtaining improved generalization even under limited training samples.

Subspace learning

Another approach for deriving compact and discriminative LBP-based feature vectors consists of applying subspace methods for learning and projecting the LBP features from the original high-dimensional space into a lower dimensional space. For instance, Chan et al. (2013) used Linear Discriminant Analysis (LDA) to project high-dimensional Multi-Scale LBP features into a discriminant space, yielding very good results. To deal with the small sample size problem of LDA, Shan et al. (2006) constructed ensemble of piecewise Fisher Discriminant Analysis (EPFDA) classifiers, each of which is designed based on one segment of the high-dimensional histogram of local Gabor binary pattern (LGBP) features. Their approach was shown to be more effective than applying LDA to high-dimensional holistic feature vector.

Tan and Triggs (2010) combined Gabor wavelets and LBP features and projected them to PCA space. Then, the Kernel Discriminative Common Vectors (KDCV) is applied to extract discriminant nonlinear compact features for face recognition. D. Zhao et al. (2007) applied Laplacian PCA (LPCA) for LBP feature selection and pointed out the superiority of LPCA over PCA and KPCA for feature selection. Hussain and Triggs (2010) exploited the complementarity of three sets of features, namely HOG, LBP, and LTP, and adopted Partial Least Squares (PLS) dimensionality reduction for selecting the most discriminative features, yielding fast and efficient visual object detector.

Nanni et al. (2012) explored using random subspace, known to work well with noise and correlated features, to train features based also on non-uniform patterns. The approach fuses classifiers (SVM) trained considering the uniform patterns and random subspace classifiers trained considering only the non-uniform patterns.

Other methods

Ren et al. (2013b) found that most existing approaches rely on a pre-defined LBP structure to extract features and that those structures can be generalized as the patterns constructed from the binarized pixel differences in a local neighborhood. Instead of using a predefined structure, they learn binarized pixel-difference patterns (BPP), casting the BPP structure discovery as a feature selection problem, which is solved via incremental minimal-redundancy-maximal-relevance (mRMR) algorithm. The method outperformed existing methods (e.g., CENTRIST (Wu & Rehg, 2011)) in two scene recognition problems.

From the observation that LBP is equivalent to the application of a fixed binary decision tree, Maturana et al. (2010) proposed a new method for learning discriminative LBP-like patterns from training data using decision tree induction algorithms. For each local image region, a binary decision tree is constructed from training data, thus obtaining an adaptive tree whose main branches are specially tuned to encode discriminative patterns in each region.. Among the drawbacks of the proposed decision tree LBP (DT-LBP) is the high cost of constructing and storage of the decision trees especially when large pixel neighborhoods are used.

### 12.3.3 Other methods inspired by LBP

LBP has also inspired the development of new effective local image descriptors related to LBP.

The local phase quantization (LPQ) descriptor is based on quantizing the Fourier transform phase in local neighborhoods (Rahtu et al., 2012). The phase can be shown to be a blur invariant property under certain commonly fulfilled conditions. In texture analysis, histograms of LPQ labels computed within local regions are used as a texture descriptor. Generation of the labels and their histograms is similar to the LBP method. Extensions of LPQ to multiple scales (Chan et al., 2013), spatiotemporal domain (Päivärinta et al., 2011) and color images have also been developed. The LPQ descriptor has received recently considerable interest in blur-invariant face recognition (Rahtu et al., 2012, Chan et al., 2013).

Lategahn et al. (2010) developed a framework which filters a texture region by a set of filters and subsequently estimates the joint probability density functions by Gaussian mixture models (GMM). Using the oriented difference filters of the LBP method (Ahonen & Pietikäinen, 2009), they showed that this method avoids the quantization errors of LBP, obtaining better results than with the basic LBP. Additional performance improvement of the GMM-based density estimator was obtained when the elementary LBP difference filters were replaced by wavelet frame transform filter banks.

Vu and Caplier (2012) proposed a feature descriptor called Pattern of Oriented Edge Magnitudes (POEM) for face recognition and image matching. First image gradient is computed and then a histogram of orientations is accumulated and assigned to the central pixel of a cell (a spatial region around the current pixel). The gradient magnitude is used to weight the contribution of each pixel. Finally, the accumulated magnitudes are encoded using the LBP operator. Descriptors based on these principles performed very well in face recognition and image matching experiments.

Sharma et al. (2012) employed Local Higher-order Statistics (LHS) of local non-binarized patterns for image description. The LHS requires neither any user specified quantization of the space of pixel patterns nor any heuristics for discarding low occupancy volumes of the space. Experiments with texture and face databases, with an SVM classifier, demonstrate very good performance. Zhang et al. (2013) proposed Local Energy Pattern (LEP) for texture classification using self-adaptive quantization thresholds. The method generates local feature vectors obtained by rectifying the responses of the 2D Gaussian-like second derivative filters, then utilizes N-nary coding quantization instead of binary one. LEP was also extended to spatiotemporal analysis.

In (He at al., 2013), texture images are first decomposed by the shearlet transform, followed by construction of local energy features. These are then quantized and encoded to be rotation invariant. The energy histograms accumulated over all decomposition levels reflect the different energy distributions. The method extracts more directional features like orientations and is robust with respect to noise. Experiments show very promising performance, especially with additive Gaussian noise.

Inspired by LBP, Maani et al. (2013) introduced a method in which local frequency components are computed by applying 1D Fourier transform on a neighboring function defined on a circle of radius R at each pixel. They observed that the low frequency components are the major constituents of the circular functions and can effectively represent textures. Three sets of features were extracted from the low frequency components, two based on the phase and one based on the magnitude. The method has advantages of a very good performance, relatively small number of features, and robustness to noise.

Wu and Rehg (2011) proposed CENTRIST, a holistic CENsus TRansform hISTogram based visual descriptor for recognizing places and scene categories. LBP-like census transform (Zabih & Woodfill, 1994) histograms are used to encode structural properties within an image and suppress detailed textural information. Constraints between neighboring pixels are utilized to capture the structural characteristic within a small image patch. In larger scales, spatial hierarchy of CENTRIST is used to catch rough geometrical information. A downside of the method is that it is not rotation-invariant.

Crosier and Griffin (2010) proposed Basic Image Features (BIF) for texture classification. BIF engineers, like LBP, a dataset-independent dictionary of local features over which textures are represented statistically. 0th, 1st and 2nd order Gaussian derivative filters are used for local description. Inspired by methods like LBP which produce binary codes, Kannala and Rahtu (2012) proposed Binarized Statistical Image Features (BSIF). BSIF computes a binary code for each pixel by linearly projecting local image patches on to a subspace, whose basis vectors are learnt from natural images via independent component analysis,

LBP has also given inspiration to recent interest on binary local feature descriptors, including BRIEF (Calonder et al., 2012), ORB and BRISK. The binary descriptors provide a comparable matching performance with the widely used interest region descriptors such as SIFT and SURF, but have very fast extraction times and very low memory requirements needed, e.g. in emerging applications using mobile devices with limited computational capabilities. Comparative evaluations of these descriptors can be found in (Heinly et al., 2012, Miksik & Mikolajczyk, 2012).

## 12.4. Spatiotemporal and other domains

### 12.4.1 Variants of spatiotemporal LBP

   High success of spatiotemporal LBP methods in various computer vision problems and applications has led to many other teams investigating the approach and several extensions and modifications of Spatiotemporal LBP have been proposed to increase its robustness and discriminative power.

Zhao and Pietikäinen (2009) extended LBP-TOP to multi-scale spatiotemporal space, with an application to facial expression recognition. AdaBoost was used to learn the principal appearance and motion, for selecting the most important expression-related features for all the classes, or between every pair of expressions. Rotation invariant variants of LBP-TOP based on histogram Fourier features, were proposed in (Zhao et al., 2012).

LTP-TOP was developed by Nanni et al. (2011). The encoding function was modified for considering both the ternary patterns and the Three Orthogonal Planes. Their experiments on 10-class Weizmann dataset obtained very good results. Weber Law Descriptor (WLD) has also been extended to spatiotemporal domain in the same way as LBP-TOP, yielding WLD-TOP for supplementing LBP-TOP in dynamic texture segmentation (Chen et al., 2013). Mattivi and Shao (2009) proposed Extended Gradient LBP-TOP for action recognition. Two modifications were made on the basis of LBP-TOP. Firstly, the computation of LBP was extended to nine slices, three for each axis. Therefore, on the XY dimension there is the original XY plane (centered in the middle of the cuboid) plus other two XY planes located at 1/4 and 3/4 of the cuboid's length. The same is done for XT and YT dimensions. Secondly, computation of LBP operator on gradient images was introduced. The gradient image contains information about the rapidity of pixel intensity changes along a specific direction, has large magnitude values at edges and it can further increment LBP operator's performances, since LBP encodes local primitives such as curved edges, spots, flat areas etc. For each cuboid, the brightness gradient is calculated along x, y and t directions, and the resulting three cuboids containing specific gradient information are summed in absolute values. Before computing the image gradients, the cuboid is slightly smoothed with a Gaussian filter in order to reduce noise. The extended LBP-TOP is then performed on the gradient cuboid. Experiments on KTH human action dataset showed the effectiveness of the method.

For dealing with 2D face recognition, Lei et al. (2011) proposed effective LBP operator on three orthogonal planes of Gabor volume (E-GV-LBP). First, the Gabor face images are formulated as a 3rd-order Gabor volume. Then LBP operator is applied on three orthogonal planes of Gabor volume respectively, named GV-LBP-TOP in short. In this way, the neighboring changes both in spatial space and during different types of Gabor faces can be encoded. Moreover, in order to reduce the computational complexity, an effective GVLBP (E-GV-LBP) descriptor was developed that describes the neighboring changes according to the central point in spatial, scale and orientation domains simultaneously for face representation.

Visual information from captured video is important for speaker identification under noisy conditions. Combination of LBP dynamic texture and EdgeMap structural features were proposed to take both motion and appearance into account (Zhao et al., 2010), providing the description ability for spatiotemporal development in speech. Spatiotemporal dynamic texture features of local binary patterns extracted from localized mouth regions are used for describing motion information in utterances, which can capture the spatial and temporal transition characteristics. Structural edge map features are extracted from the image frames for representing appearance characteristics. Combination of dynamic texture and structural features takes both motion and appearance together into account, providing the description ability for spatiotemporal development in speech. In the experiments on BANCA and XM2VTS databases, the proposed method obtained promising recognition results comparing to the other features.

Goswami et al. (2010) proposed a novel approach to ordinal contrast measurement called Local Ordinal Contrast Patterns (LOCP). Instead of computing the ordinal contrast with respect to any fixed value such as that at the center pixel or the average intensity value, it computes the pairwise ordinal contrasts for the chain of pixels representing the circular neighborhoods starting from the center pixel.

Then it was extended for dynamic texture analysis by extracting the LOCP in three orthonormal planes to generate LOCP-TOP. Together with LDA, it's performance of mouth-region biometrics in the XM2VTS database received good results.

Spatial representation of LPQ was extended to a dynamic texture descriptor called the Volume Local Phase Quantization (VLPQ) by Päivärinta et al. (2011). The local Fourier transform is computed by 1D convolutions for each dimension in a 3D volume. The data achieved is compressed to a smaller dimension before a scalar quantization procedure. Finally, a histogram of all codewords from dynamic texture is formed.

Huang et al. (2012) proposed to use spatiotemporal monogenic binary patterns to describe the appearance and motion information of the dynamic sequences. Firstly, they use monogenic signals analysis to extract the magnitude, the real picture and the imaginary picture of the orientation of each frame, since the magnitude can provide much appearance information, and the orientation can provide complementary information. Secondly, the phase-quadrant encoding method and the local bit exclusive operator are utilized to encode the real and imaginary pictures from orientation in three orthogonal planes, and the local binary pattern operator is used to capture the texture and motion information from the magnitude through three orthogonal planes. Finally, both the concatenation method and multiple kernel learning method are exploited to handle the feature fusion. The experimental results on the Extended Cohn-Kanade and Oulu-CASIA facial expression databases demonstrated a state-of-the-art performance.

Ruiz-Hernandez and Pietikäinen (2013) proposed a method for encoding LBPs using a re-parametrization (RP) of the second local order Gaussian Jet. The information provided by RP generates robust and reliable histograms, and is thus suitable for different facial analysis tasks. The proposed method has two main processes: the RP process, which is used to compute needed parameters in a video sequence, and the encoding process, which combines the textural information provided by the LBP and the robustness of the re-parametrization. They showed that this approach can be used for recognizing facial micro-expressions from videos, obtaining competitive performance on two different datasets.

Video texture synthesis is the process of providing a continuous and infinitely varying stream of frames. Guo et al. (2013) proposed a variant of LBP-TOP called Multi-frame LBP-TOP to find the most appropriate matching pairs of frames for video texture synthesis. To achieve seamless synthesis results, a diffeomorphic growth model was applied to matching frames identified. The proposed approach has potential other applications in motion interpolation for videos and dynamic texture retrieval.

Inspired by spatiotemporal LBP, the Haar-like features were extended to represent the dynamic characteristic of facial expression (Yang et al., 2009). The dynamic Haar-like features were built by two steps: (1) thousands of Haar-like features are extracted in each frame and (2) the same Haar-like features in the consecutive frames are combined as the dynamic features as Fig. 12.7 shows.

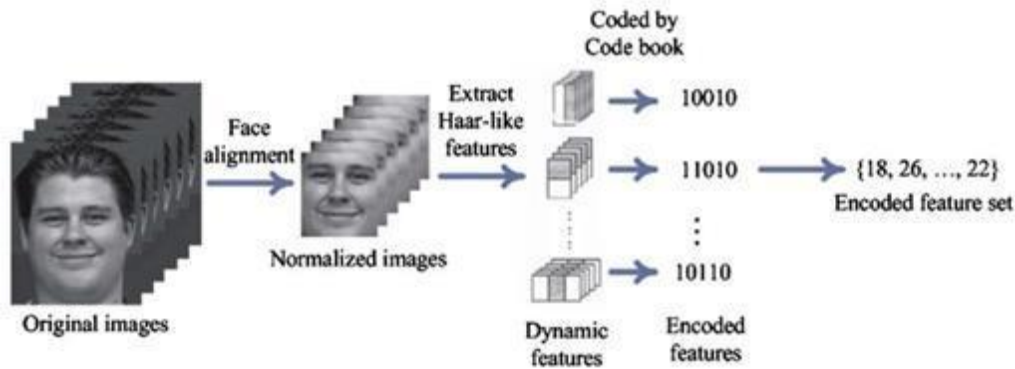

Fig. 12.7. Dynamic Haar-like features.

With the increasing amount of surveillance data, segmentation of moving objects in the compressed domain is receiving growing attention. Yang et al. (2012) proposed a method, in which the motion vectors are first accumulated and filtered to have reliable motion information. Then, spatiotemporal LBP features of motion vectors are extracted to obtain coarse and initial object regions in the H.264 compression domain. Final region refinement is done according to the distribution of DCT coefficients.

A local spatiotemporal directional descriptor was proposed for speaker identification by analyzing mouth movements (Zhao & Pietikäinen, 2013). The movements of mouth regions are described using LBPs and intensity contrast from six directions in three orthogonal planes. In addition, besides sign features, magnitude information encoded as weight for the bins with the same sign value was developed to improve the discriminative ability. Moreover, decorrelation is exploited to remove the redundancy of features. Experimental results on the challenging XM2VTS database showed the effectiveness.

*12.4.2 LBP in other domains*

Analysis of depth and 4D images

Research on LBP variants for depth (range) images has focused on applications in face analysis. In most of the early papers, the depth values were first interpolated onto a regular grid, and then a more or less ordinary 2D LBP approach was applied to the intensity (depth) values (Huang et al., 2006).

Huynh et al. (2013) improved these by considering orientation differences in LBP computations, obtaining eight $LBP_{8,1}$ based orientations of the depth differences in a local 3 x 3 neighborhood. Eight depth difference histograms are obtained for the given region and concatenated to form an oriented histogram of depth differences. Sandbach et al. (2012) extended the LBP methodology to 3D face data by utilizing surface orientation information to detect facial action units (AU). Local Normal Binary Patterns (LNBP) employ the normals of the triangular polygons that form the 3D face mesh to encode the shape of the mesh at each point. This is related to the gradient of a 2D image, providing a richer source of local shape than the intensity alone. Bayramoglu et al. (2013) also used surface orientation information in their method for facial action unit detection. To get a short feature vector without degrading the performance, they used the center-symmetric LBP (CS-LBP) approach with 3D surface orientation information in computing the CS-3DLBP operator. In (Tang et al., 2013), with the help of extracted keypoints the 3D face is divided into a mesh. LBP is computed using eight neighboring vertices of a current vertex. The depth and normal information of each vertex are extracted separately and encoded by LBP. The depth information describes the concavo-convex attribute, and the normal information indicates the face curvature variety.

Due to the progress of depth sensor technology, the analysis of 4D images (i.e., depth images in motion) is an emerging research topic, for example in facial expression recognition. Fang et al. (2011) adopted the spatiotemporal LBP-TOP operator for 4D facial expression recognition. A robust approach for registering consecutive frames of 4D data is first applied and the resulting "geometry" images are considered as frames in 2D videos. LBP-TOP descriptors are computed on the difference between a frame and the first frame of the sequence, using the idea of so called flow image representing the deformation vectors in a subsequent frame w.r.t. the first frame. Promising results are obtained for the BU-4DFE facial expression database. Reale et al. (2013) used LBP-TOP as a comparative method in their paper on 4D spatiotemporal "Nebula" feature. The LBP-TOP approach, applied to 2D and depth images, was in many ways similar to that of (Fang et al., 2011), but did not use their alignment method and their tests on a flow image. The results for the Nebula feature were best, but also LBP-TOP performed reasonably well considering its simple way of implementation in this study.

Analysis of 3D volume images

 Extension of LBP to 3D volume images is challenging. A circle in 2D translates to a sphere in 3D, and equidistant sampling on a sphere is not so trivial. The notion of ordering is also lost in 3D due to the dimensionality (Banerjee et al., 2013).

Fehr and Burkhardt (2008) extended the original LBP from 2D images to 3D volume data, achieving a full rotation invariance. For each LBP computation, correlation between the values of all points on the neighborhood sphere with radius R and the weight factor which is a volume representation in an arbitrary but fixed order binomial factors is performed in the spherical harmonic domain. Rotation invariance is obtained from the computation of the minimum over all angles. Another method for rotationally invariant 3D LBP using spectral harmonic decomposition was proposed by Banerjee et al. (2013). Unlike the original approach, the invariance is constructed implicitly, without considering all possible combinations of the pattern. Experiments were carried out with phantom data and a clinical liver CTA data.

Morgado et al. (2013) proposed an approach able to closely replicate in 3D and without any approximation both uniformity and rotation invariance concepts originally proposed for the 2D setting. Feature selection is done using correlation coefficients to quantify the relevance of each individual feature, and the SVM is used as the learning machine. The experiments demonstrate that the proposed method is able to enhance the diagnostic system and that the texture of the FDG-PET scans contains distinctive information about the presence of both Alzheimer's disease and mild cognitive impairment. Burner et al. (2011) consider sign, contrast and intensity information in computing texture bags at multiple scales for 3D volumes. Instead of using histograms to match entire images, they search for regions that have similar local appearance. This is necessary to cope with the variability encountered in medical imaging data, and the comparably subtle effects of disease. Based on the similarity of the texture structure of local regions, images are being ranked.

LBP in 1D signal analysis

Analysis of 1D signals is an emerging and potentially very important application area for LBP. For example, speech systems such as hearing aids require fast and inexpensive signal processing. Chatlani et al. (2010) proposed a straightforward simplification of the ordinary LBP to 1D signals, with a preliminary application to simple signal segmentation and voice activity detection (VAD) to estimate periods of speech and non-speech. Later, Zhu et al. (2012) used LBP-based VAD in HMM-based speech recognition. Speech is first denoised by Adaptive Empirical Model Decomposition (AEMD) and processed with LBP-based VAD, in which 1D LBP is used to find start and end points of voiced speech segment so that it is distinguished from noise, unvoiced or mute segments. In another application example, 1D LBP is applied for bone texture classification by Houam et al. (2012). Global texture information is characterized by image projections, and local information is extracted from these using 1D LBP.

Another interesting direction for 1D signal analysis is to first derive a 2D representation of the signal and then apply spatial domain LBP to this representation. Lazic and Aarabi (2008) applied successfully 2D LBP for detecting spoken terms from visual spectrogram representation derived from the audio signal. A similar approach was used by Costa et al. (2012) for classifying music genre from visual spectrograms derived from the audio signal. Esfahanian et al. (2013), on the other hand, borrowing principles of facial image analysis (Ahonen et al., 2006), divided the visual spectrograms first into non-overlapping regions and then used concatenated histograms to classify dolphin calls.

## 12.5. Future challenges

Due to the considerable effort on spatial domain variants, the future research should focus more on new challenges.

A great majority of the research on dense texture descriptors has been based on the assumption that there are no significant view or scale variations in the scene or objects to be analyzed. In practice, these variations are very common and may include self-occlusion when an object is imaged from different views. An obvious but impractical solution would be to train the recognition system with samples viewed from a large number of different positions, and then derive a set of representative models for each class (Pietikäinen et al., 2011). Moore and Bowden (2011) showed that a larger-sized operator (Gabor LBP or Multiscale LBP) works better than small-sized operators in multi-view facial expression recognition. It would very be useful to develop new approaches that are designed for handling effectively problems of view and scale variation.

For a large number of applications an ability to analyze small sample sizes at high speed is vital, including face analysis, interest region description, segmentation, background subtraction, and tracking. This means that a compact region description with a short feature vector is needed. Many of the proposed descriptors would fail in this respect. It is important to evaluate the performance of a new descriptor also with smaller sample sizes, as was done e.g. in (Ylioinas et al., 2013b) by cropping 41 x 41 patches from the original 200 x 200 pixel CUReT textures.

There is also need for more research on developing approaches which are robust to lighting variations. The use of Gabor filtering (W. Zhang et al., 2005) or a preprocessing by Tan and Triggs (2010) are examples of the state of the art for face recognition, but these approaches also tend to smooth small details from the images. Some of the recent works have successfully used image gradient orientation computation prior to feature extraction, as gradient orientations are known to be less sensitive to lighting variations than the original images.

Most of the past research has focused on improving the robustness of LBP for some specific tasks, like preprocessing prior to LBP computation (e.g. Gabor filtering, gradient computation), using different methods for encoding (e.g. LTP), reducing the number of neighbors in multi-scale analysis (e.g. LQP), processing LBP codes in different ways (e.g. LBP-HF). A future direction would be to consider different stages of feature computation together to optimize the performance. A good example of this direction is by Lei et al. (2014), who obtained excellent results in face recognition by combining preprocessing by learning-based linear filtering, soft-sampling to find best set of neighbors for LBP computation, and clustering to find optimal way for encoding.

Combining different types of complementary local operators is another way to go ahead, as e.g. the recent works on human detection and face recognition show. Approaches combining the strengths of LBP and HOG (or SIFT), for example, have led to increased performance.

Research on spatiotemporal LBP variants has not been as active as in spatial domain. One would expect to see much more research in this area. Different effective ways of obtaining information in temporal domain combined with novel ideas proposed for analyzing images in spatial domain could be one way to go ahead. The robustness of spatiotemporal operators to different variations is largely unexplored. One example of this kind of work is by Zhao et al. (2012), who demonstrated that their spatiotemporal LBP-HF features are robust with respect to view variations.

The use of LBP for analyzing 1D signals has not been much studied, but has potential for a large number of novel applications, as the recent examples presented above demonstrate. In these works quite primitive LBP-based analysis was used both for 1D signal analysis and for 2D analysis of spectrograms derived from 1D signals. Borrowing ideas from the recent developments in spatial domain LBP variants could lead to much more powerful methods. The results on analyzing 2D spectrograms also suggests that LBP variants could be very useful in a wide variety of applications of exploratory data analysis, in which relations between neighboring data elements should be considered.

With the introduction of Kinect and emerging more accurate sensors for depth sensing, the interest on processing depth images (3D) and depth image sequences (4D) has been growing, with applications, e.g. in action and gesture recognition, face recognition, and recognition of facial action units. Some LBP variants for depth images have been developed especially for face analysis, but research on 4D data is largely unexplored. There is also much space for new developments in 3D volume image analysis widely used in medicine. Simple 3D variants of ordinary LBP have provided very promising improvement, but much more could be possible with more sophistigated solutions.

The databases used for comparing texture descriptors should be reconsidered. A good practice in some recent papers is to use both texture and face databases, as well as datasets from other application domains. For 2D faces there are many good options available like Labeled Faces in the Wild (LFW) and Face Recognition Grand Challenge (FRGC v2.0), representing different challenges. A problem with most of the used texture databases is that the best methods currently obtain over 95% accuracy. KTH-TIPS-2 represents a more difficult problem in this respect. It would be valuable to have extensive benchmarks with challenging databases for the most promising variants, as was done for texture retrieval in (Doshi & Schaefer, 2012).

12.6. Conclusions

Due to its advantages, i.e., flexibility, invariance to monotonic gray-level changes, and computational simplicity, LBP is a very powerful descriptor to represent local structures in images. A large number of variants have been designed aiming to obtain improved performance and/or robustness in one or more aspects of the original LBP. We have divided the extensions and modifications into ten categories and introduced some representative variants in each of them.

A number of new effective local image descriptors, including Local Phase Quantization (LPQ), Patterns of Oriented Edge Magnitudes (POEM) and Local Energy Patterns (LEP), have also been inspired by LBP. They provide the ability, e.g. to deal with blur or noise, and can be jointly used with LBP to complement each other.

Even though it is commonly agreed that multi-scale analysis and joint use of complementary descriptors can improve the performance, a downside is the large dimensionality of the produced feature vector. To obtain a small set of the most discriminative LBP-based features, different feature selection and learning strategies were also discussed. Extensions of LBP to spatiotemporal domain extend the applicability of LBP from static images to dynamic video sequences. Different variants of the original spatiotemporal LBP were introduced.

Moreover, the future challenges of LBP were discussed. Among these are an improved robustness to view, scale and lighting variations, optimization of different stages of feature computation, and combinations of different descriptors. There is also need for further research on problems such as spatiotemporal LBPs and analysis of 1D signals, depth and 3D volume images, and 4D depth image sequences.